\documentclass[runningheads]{llncs}

\usepackage{eccv}

\usepackage{eccvabbrv}

\usepackage{graphicx}
\usepackage{epsfig}
\usepackage{booktabs}
\usepackage[accsupp]{axessibility}  
\usepackage{colortbl}
\usepackage{amsmath}
\usepackage{longtable}
\usepackage{caption}
\usepackage{multirow}
\usepackage{color,xcolor}

\definecolor{tabfirst}{rgb}{1, 0.7, 0.7} 
\definecolor{tabsecond}{rgb}{1, 0.85, 0.7} 
\definecolor{tabthird}{rgb}{1, 1, 0.7} 
\definecolor{lightgray}{rgb}{0.9, 0.9, 0.9} 
\definecolor{LightCyan}{rgb}{0.88,1,1}

\newlength\savewidth\newcommand\shline{\noalign{\global\savewidth\arrayrulewidth
  \global\arrayrulewidth 1.25pt}\hline\noalign{\global\arrayrulewidth\savewidth}}

\newcommand{\repeatthanks}{\textsuperscript{\thefootnote}}
\usepackage{graphicx}
\usepackage{caption}
\usepackage{subcaption}
\usepackage{array}
\usepackage{colortbl}
\usepackage{multirow}
\usepackage{float}
\usepackage{tabularx}

\usepackage{hyperref}

\usepackage{orcidlink}

\begin{document}

\title{
\makebox[5pt][l]{\raisebox{-0.7ex}{\includegraphics[height=36pt,trim={0cm 0cm 0cm 0cm},clip]{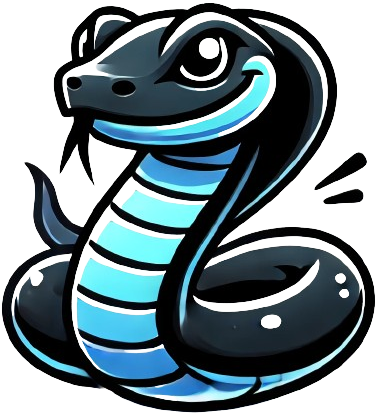}}}\hspace{25pt}
VideoMamba: Spatio-Temporal Selective State Space Model
}

\titlerunning{VideoMamba: Spatio-Temporal Selective State Space Model}

\author{Jinyoung Park\thanks{Equal contribution.}\orcidlink{0000-0001-7129-2141} \and
Hee-Seon Kim\repeatthanks\orcidlink{0009-0009-4006-6112} \and
Kangwook Ko\repeatthanks\orcidlink{0000-0002-5101-9751} \and
Minbeom Kim\orcidlink{0009-0007-0300-0988} \and
Changick Kim\orcidlink{0000-0001-9323-8488}}

\authorrunning{J.~Park et al.}

\institute{Korea Advanced Institute of Science and Technology\\
\email{\{jinyoungpark, hskim98, kw.ko, alsqja1754, changick\}@kaist.ac.kr}\\
\url{https://github.com/jinyjelly/VideoMamba} 
}
\maketitle

\begin{abstract}
We introduce VideoMamba, a novel adaptation of the pure Mamba architecture, specifically designed for video recognition. 
Unlike transformers that rely on self-attention mechanisms leading to high computational costs by quadratic complexity, VideoMamba leverages Mamba's linear complexity and selective SSM mechanism for more efficient processing.
The proposed Spatio-Temporal Forward and Backward SSM allows the model to effectively capture the complex relationship between non-sequential spatial and sequential temporal information in video.
Consequently, VideoMamba is not only resource-efficient but also effective in capturing long-range dependency in videos, demonstrated by competitive performance and outstanding efficiency on a variety of video understanding benchmarks.
Our work highlights the potential of VideoMamba as a powerful tool for video understanding, offering a simple yet effective baseline for future research in video analysis. 

\end{abstract}

\section{Introduction}
\label{sec:intro}
In the field of natural language processing, Transformer \cite{vaswani2017attention} has demonstrated remarkable performance. 
After the success of Vision Transformer \cite{dosovitskiy2020image}, transformers began to be utilized across various computer vision problems, surpassing the performance of previous CNN-based methods \cite{carion2020end, chen2021pre, wang2021transformer, zheng2021rethinking, arnab2021vivit}.
However, the core operator of Transformers, self-attention, presents a challenge due to its quadratic complexity.
This becomes particularly problematic in video recognition tasks that require the processing of multiple frames, limiting the applicability in resource-constrained environments.

\begin{figure}[h!]
    \centering
         \includegraphics[width=\linewidth
         ,trim={0cm 0.8cm 0.7cm 0.7cm},clip]
    {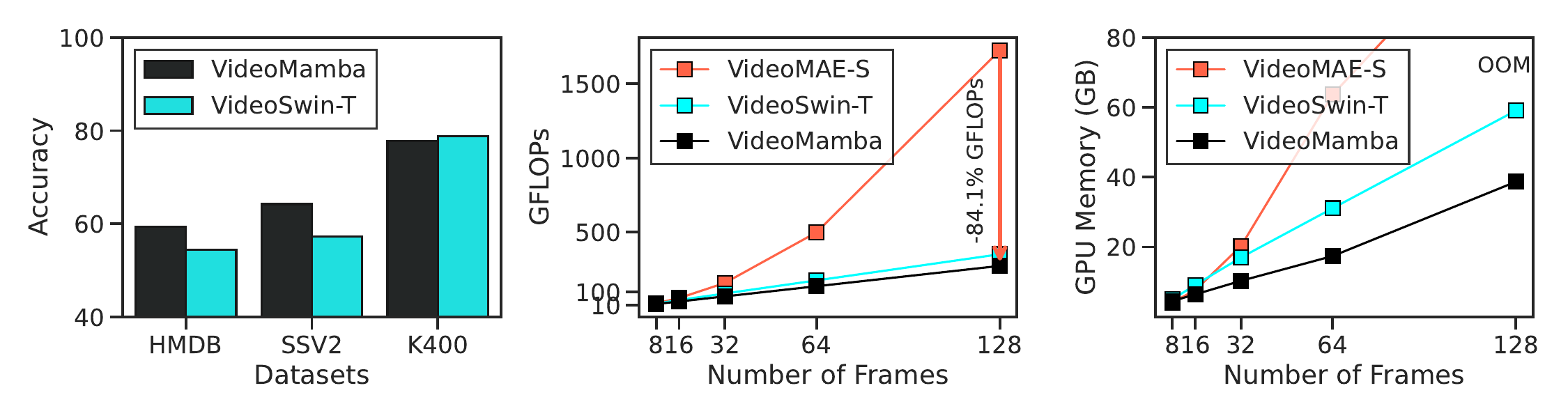}
        \caption{\textbf{Performance and efficiency comparisons among ImageNet-1K pretrained video models.} 
VideoMamba shows superior or comparable performance to VideoSwin-T \cite{liu2022video}, while having clear advantage in terms of reduced GFLOPs and memory consumption compared to VideoSwin-T \cite{liu2022video} and VideoMAE-S \cite{wang2023videomae}.}
    \label{fig:fig1}
\end{figure}

Recently, Mamba \cite{gu2023mamba} is introduced, offering an appealing alternative to this problem.
Based on structured Selective State Space Models (SSMs), Mamba employs a selective scan mechanism that dynamically adjusts parameters based on the input.
With the selective scan mechanism and hardware-aware algorithm, Mamba effectively captures long-range dependency with contextual awareness, while maintaining linear complexity.
Consequentially, Mamba achieves superior performance compared to transformers in various 1D sequence modeling problems, such as language modeling, audio processing, and genomics.
Efforts to apply this innovative mechanism to a range of vision tasks, such as image classification and segmentation, are currently underway \cite{zhu2024vision, liu2024vmamba, liang2024pointmamba}.

Inspired by the success of Mamba, we propose \textbf{VideoMamba}, the first video model to offer a comprehensive analysis of adapting the pure Mamba architecture for video tasks.
Our video model not only demands lower computational resources but also delivers competitive performance compared to its transformer counterparts of similar size.  
While the Mamba model's linear complexity and ability to capture long-range dependency are well-suited for video applications, representing the spatio-temporal information of videos into a 1D sequence presents a significant challenge.
To effectively address the issue of handling non-sequential spatial information, recently introduced variants of the mamba models in vision \cite{zhu2024vision, li2024mamba,liu2024vmamba} adopt bidirectional scanning methods.
Building on this foundation, VideoMamba also adopts the bidirectional scanning strategy for video data, presenting a pioneering study on expanding vision models to video applications and making the pre-trained model available.
However, video data introduces a 
more challenging scenario due to the interlinked relationship between non-sequential spatial information (e.g., the position and posture of a person in specific frames) and sequential temporal information (e.g., changes in a person's actions over frames). 

\begin{figure}[t]
    \centering
     \includegraphics[width=\textwidth,trim={0cm 0cm 0cm 0cm},clip]{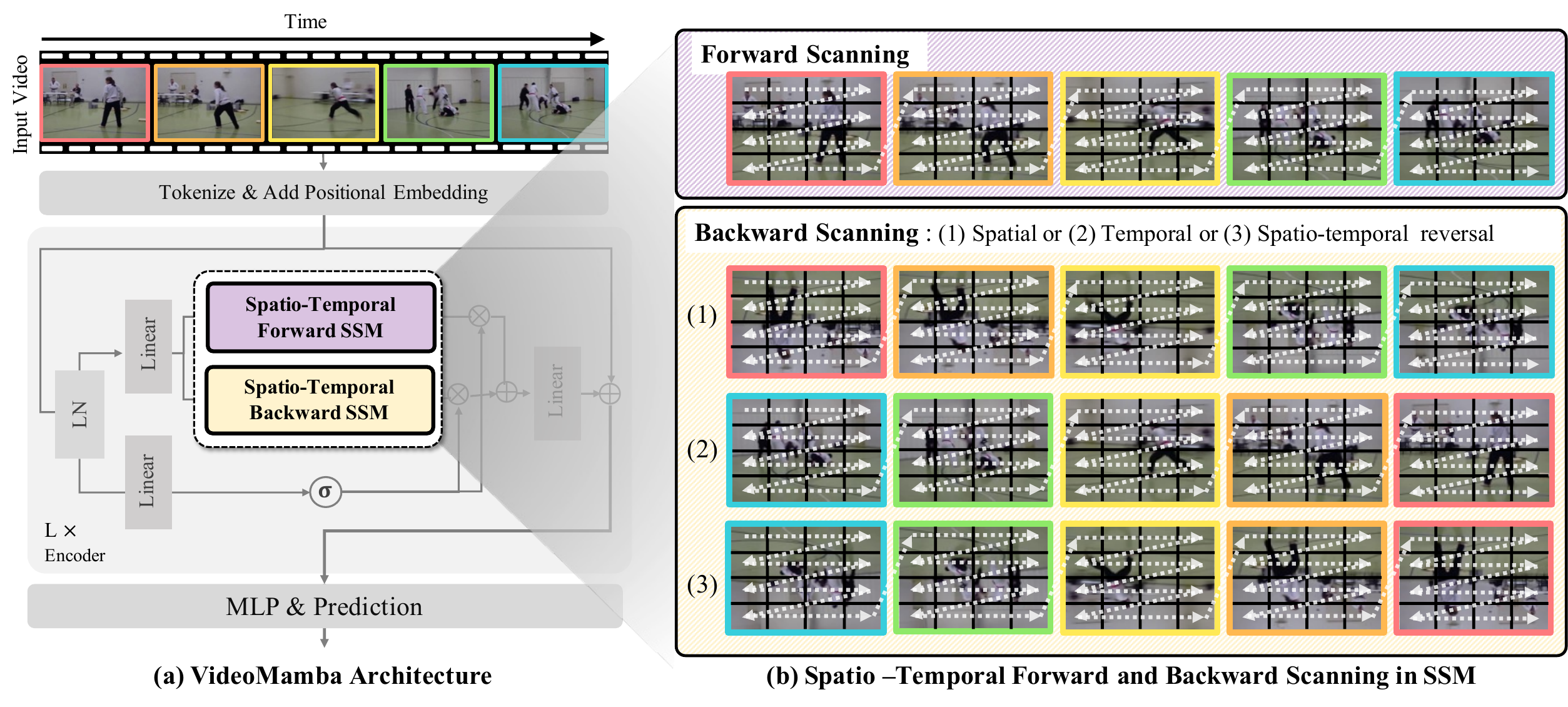}
    \caption{\textbf{Comprehensive View of VideoMamba's Framework.}
(a) Architecture of VideoMamba. This includes the initial tokenization of video frames, addition of positional embeddings, and processing through encoder blocks that utilize proposed Spatio-Temporal Forward and Backward SSMs for thorough video analysis.
(b) Process of Spatio-Temporal Forward and Backward Scanning within the SSMs, with white dashed arrows indicating the scanning direction of video tokens.}
\vspace{-3mm}
    \label{fig:overview}
\end{figure}

Our carefully designed experiments aim to tackle the inherent complexity of processing spatio-temporal information.
To extend the bidirectional SSM module effectively, we develop it into \textit{Spatio-Temporal Forward and Backward SSM} modules.
Focusing on defining the backward scanning direction, we explore the effects of spatio-temporal scanning among spatial, temporal, and spatio-temporal reversal.
Through extensive experiments and ablation studies,
we provide insights and examine various design choices and training recipes. 
Additionally, we analyze how the VideoMamba model responds to temporal consistency, investigating whether the model merely treats video as a bundle of images or not. 
Moreover, an analysis of one of Mamba's key components, Delta, offers valuable insights into how the model captures and processes spatio-temporal information.

VideoMamba has demonstrated its competitiveness across various video benchmarks. It exhibits either superior performance in HMDB 51 \cite{kuehne2011hmdb} and Something-Something V2 \cite{goyal2017something}, or comparable performance in Kinetics-400 \cite{kay2017kinetics} compared to other video models. 
Its efficiency stands out, as 
VideoMamba's outstanding reduction in GFLOPs and memory usage is highlighted in Fig. \ref{fig:fig1}.

The contributions of this paper are: 
\begin{itemize}
   \item We introduce VideoMamba, a pioneering study of a simple yet efficient pure-Mamba-based model, highlighting its potential for future advancements in video recognition.
   \item Through Spatio-Temporal Forward and Backward SSM, we tackle the unique challenge of integrating non-sequential spatial with sequential temporal information in video processing.
   \item We demonstrate VideoMamba's competitive performance and its reduced computational demands compared to conventional transformer models.
   \item Our extensive experiments and analysis underscore the Mamba model's strength as an effective tool for video processing.
\end{itemize}

\section{Related work}
\label{sec:relwork}
\subsection{Video Understanding}
Traditionally, CNNs \cite{tran2015learning, ji20123d, hara2017learning, qiu2017learning, feichtenhofer2019slowfast, kondratyuk2021movinets,lin2019tsm} have been the standard backbone architectures in video modeling.
These CNNs leverage 3D convolutions \cite{tran2015learning, feichtenhofer2019slowfast, carreira2017quo,benaim2020speednet,qian2021spatiotemporal} or factorize spatial and temporal convolutions\cite{qiu2017learning, xie2018rethinking, tran2018closer,han2020memory} for efficiency. 
Before the rise of pure Transformer-based models, the integration of attention mechanisms\cite{wang2018non, kozlov2020lightweight,gavrilyuk2020actor}, renowned for their efficiency in capturing long-range dependencies within data, into CNN frameworks achieved promising results.
The success of non-local networks\cite{wang2018non} and hybrid models\cite{yin2020disentangled, wu2019long} led to the development of pure Transformer architectures specifically designed for video recognition \cite{akbari2021vatt, arnab2021vivit, liu2022video, wu2021towards, bertasius2021space, fan2021multiscale, wang2021pyramid, neimark2021video}. 
However, a significant challenge with pure Transformer architectures lies in their computational complexity for lengthy videos. 
This stems from the quadratic complexity of the attention mechanism as the sequence length increases\cite{vaswani2017attention}. 
To address this issue, several approaches have been proposed, including factorization techniques \cite{arnab2021vivit,bertasius2021space}, windowing mechanisms\cite{liu2022video}, and masked token reconstruction \cite{tong2022videomae,feichtenhofer2022masked,wang2023videomae}. 
These methods achieve promising results in handling long sequences while capturing spatial and temporal information concurrently. 
While pure Transformers demonstrate remarkable capabilities, the inherent quadratic complexity with sequence length remains an open challenge that necessitates further exploration for efficient video understanding.

\subsection{State Space Models}
Derived from the classical state space models (SSMs)\cite{kalman1960new}, structured SSMs (S4)\cite{gu2021efficiently, gu2021combining} have emerged as promising frameworks in modeling long-distance sequences with a linear increase in computational cost.
S4's success sparked a wave of research, resulting in diverse S4-inspired models that capture long-range dependencies in sequential data\cite{gupta2022diagonal,smith2022simplified}. These variants achieve competitive performance on various tasks\cite{nguyen2022s4nd, islam2022long, wang2023selective}.
S4's strength lies in its adherence to Linear Time Invariance (LTI), ensuring consistent output for identical inputs irrespective of their temporal application. 
While LTI systems present several advantages, they also introduce limitations, particularly in handling time-variant dynamics. 
The constraint that the internal state transition matrix remains constant across the sequence restricts the model's ability to adapt to evolving content, limiting its application in scenarios requiring content-based reasoning.

Recently, Mamba\cite{gu2023mamba} addressed the limitations by introducing a selective state-space model that dynamically adjusts its parameters based on the input sequence. 
This flexibility allows Mamba to perform context-dependent reasoning, significantly enhancing its applicability across various domains from language and speech \cite{gu2023mamba} to complex visual data \cite{zhu2024vision,liu2024vmamba,li2024mamba,liang2024pointmamba}. 
While concurrent work \cite{li2024mamba} shows the potential for extension into multi-dimensions, its particular application in video recognition tasks remains unexplored. 

In this work, we propose VideoMamba, an extension of the Mamba model specifically designed for video understanding tasks. It leverages Mamba's capabilities to enhance long-range modeling for video data.
This approach builds upon the recent advancements in SSMs for video tasks. Unlike prior work, S4ND\cite{nguyen2022s4nd}, which struggled to capture input-dependent information, VideoMamba incorporates a selective scan mechanism to address this limitation.

\section{Preliminary}
\subsection{State Space Model}

State space models are linear time-invariant systems that map 1-D input sequence to 1-D output sequence through a latent state. Mathematically, these models can be expressed as simple ordinary differential equations (ODE) as follows:
\begin{equation}
\label{eqn:ssm_cont}
\begin{split}
h'(t)={A}h(t)+{B}x(t),\\
y(t)={C}h(t)+{D}x(t),
\end{split}
\end{equation}
where $x(t) \in \mathbb{R}$ is a continuous input sequence, $y(t) \in \mathbb{R}$ is an output sequence, and $h(t) \in \mathbb{R}^{N}$ denotes a latent state.

To enable the processing of discrete signals in Equations \ref{eqn:ssm_cont}, a discretization process is necessary. The most commonly used method for discretization is the zero-order hold (ZOH) technique, which transforms parameters $A$ and $B$ for continuous signals into parameters $\bar{A}$ and $\bar{B}$ for discrete signals through the step size parameter $\Delta$. The parameter discretization process is formulated as follows:
\begin{equation}
\label{eqn:discretization}
\begin{split}
&\bar{A} = \exp(\Delta A),\\
&\bar{B} = (\exp(\Delta A) - I)(\Delta A)^{-1}B ,\\ 
&\bar{C} = C.\\
\end{split}
\end{equation}

After the discretization, the continuous system in Equations \ref{eqn:ssm_cont} can be written in discrete form as
\begin{equation}
\label{eqn:ssm_discrete}
\begin{split}
&h_k = \bar{A}h_{k-1} + \bar{B}x_k, \\ 
&y_k = \bar{C}h_k,\\
\end{split}
\end{equation}
where $x_{k}$ and $y_{k}$ are discrete input and output signals. 

\subsection{Selective SSM}
While SSMs have demonstrated outstanding performance in various tasks with sequential data, as previously mentioned, they suffer from the inherent limitation of being an LTI system. In other words, with parameters $A$, $B$, $C$, and $\Delta$ remaining constant across all time steps, the model's computations are independent of the input, making it challenging to address problems requiring contextual awareness. Recently, to overcome these limitations, Mamba \cite{gu2023mamba} is proposed, leveraging a \emph{selective scan} mechanism. In Mamba, model parameters such as $B$, $C$, and $\Delta$ are dynamically determined based on the input, enabling the model to understand the context of input sequences. 
We adopt this selective SSM as the core operator in our proposed model.

The key parameter of selective SSM, $\Delta$, acts as a gating mechanism that controls the influence of specific elements within the state transition matrices (A, B, and C), and these matrices determine how the model's hidden state evolves over time.
More specifically, as stated in the original paper \cite{gu2023mamba}, large $\Delta$ represents that the hidden state is ignored and current input is highlighted, and small $\Delta$ represents that current input is ignored. 
In the context of video understanding, $\Delta$ enables VideoMamba to selectively focus on relevant aspects of the hidden state for updates, aiding in context-dependent reasoning.
More analysis and visualization of $\Delta$ is given in Sec. \ref{sec:delta}.

\setlength{\belowcaptionskip}{-5pt} 
\begin{figure}[t!]
    \centering
     \includegraphics[width=1.0\textwidth,trim={0cm 0cm 0cm 0cm},clip]{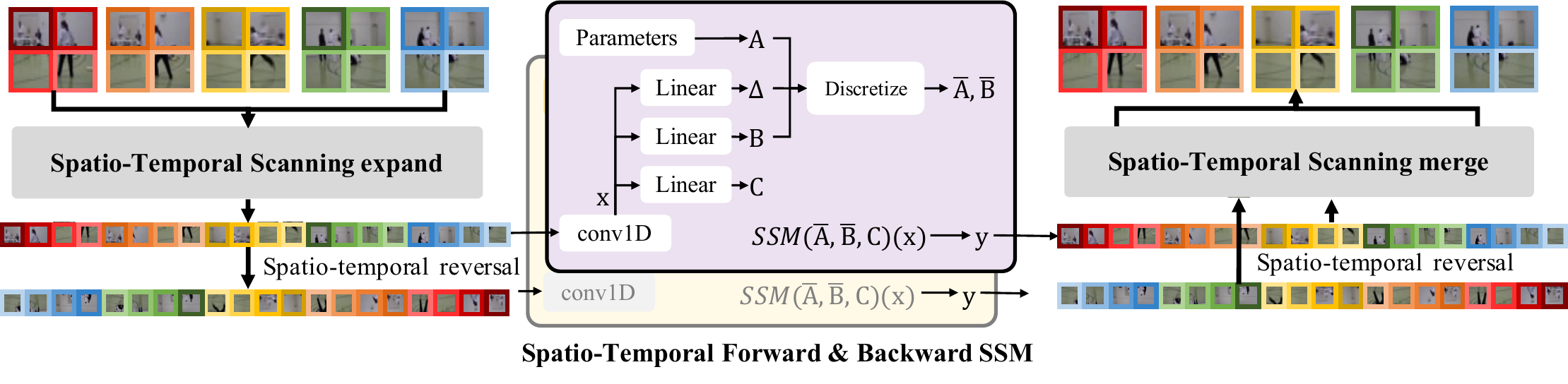}
    \caption{\textbf{Details of proposed Spatio-Temporal SSM.}
The figure illustrates the external and internal operations of the Spatio-Temporal Forward \& Backward SSM. 
Here, the backward scanning method represents Spatio-temporal reversal.}
    \label{fig:block}
\end{figure}

\section{VideoMamba} 
\label{section:VideoMamba}
The overall architecture of the proposed VideoMamba is presented in Fig. \ref{fig:overview}. First, the sampled video clip is transformed into video tokens via a video tokenizer (Sec. \ref{section:tokenizer}). These video tokens are then added with positional embeddings (Sec. \ref{section:embedding}), which contain positional information. These tokens, along with the class token, form the input for the model.
The input tokens pass through $L$ layers of the VideoMamba encoder (Sec. \ref{section:encoder}), where spatio-temporal forward and backward SSM (Sec. \ref{section:scanning}) is applied.
Finally, after passing through the last layer, the class token is normalized and fed to a video classification head, which consists of a single MLP layer, to generate the final prediction of the model. To overcome the difficulties in model training due to the relatively small size of video datasets, we initialize our video models using pre-trained image models, capitalizing on the strengths offered by large-scale image datasets. The detailed explanations for each component are provided in the following subsections.
\subsection{Video Tokenizer}
\label{section:tokenizer}
Video tokenizer maps the sampled video clip $V \in \mathbb{R}^{T \times H \times W \times C}$ into a sequence of video tokens $z = [z_{1}, z_{2}, \cdots, z_{n_{t} \times n_{h} \times n_{w}}]$, where  $T$, $H$, $W$ and $C$ represent the temporal length, height, width and channel of the video, respectively and each video token $z_{i} \in \mathbb{R}^{d}$ is a $d$ dimensional feature embedding.
First, we divide the video clip into non-overlapping tubelets with size of $s_{t} \times s_{h} \times s_{w}$, resulting in $n_{t} \cdot n_{h} \cdot n_{w}$  number of tubelets, where $n_{t} = \lfloor \frac{T}{s_{t}} \rfloor$, $n_{h} = \lfloor \frac{H}{s_{h}} \rfloor$, $n_{w} = \lfloor \frac{W}{s_{w}} \rfloor$.
 
Then, we utilize 3D convolutional layer to extract
video tokens from each tubelet. 
The video tokenizer is initialized from pretrained image model, by inflating the 2D convolutional layer to the 3D convolutional layer by expanding the weight tensor to the temporal axis and averaging.

\subsection{Positional Embedding} 
\label{section:embedding}
Positional embedding was first introduced in Transformer \cite{vaswani2017attention}, and is widely used in transformer-based vision models \cite{dosovitskiy2020image, arnab2021vivit}. However, positional embedding is not employed in SSMs \cite{gu2021efficiently, nguyen2022s4nd, gu2023mamba}, since the inherent recurrent nature of SSM negates the necessity for positional information. Even though, considering the characteristics of video, the use of positional embedding offers the advantage of supplementing the model with additional spatio-temporal information for each token. Therefore, as an SSM-based video understanding model, to explore the effect of using positional embedding, we consider several options for positional embedding: 
\textbf{(1) not using} any positional embedding, \textbf{(2) Sinusoidal} positional embedding and \textbf{(3) learnable} positional embedding. For options (2) and (3), the positional embedding $P \in \mathbb{R}^{n_{t} \cdot n_{h} \cdot n_{w} \times d}$ is added to input tokens $z$.

When using learnable positional embedding, we can initialize the learnable positional embedding $P$ using learned positional embedding from the pretrained image model $P_{image} \in \mathbb{R}^{n_{h} \cdot n_{w} \times d}$, which can be regarded as case when $n_{t}=1$. We consider several initialization methods: (3-1) expanding the learned positional embedding $P_{image}$ to the temporal axis by replicating it $n_{t}$ times, (3-2) interpolating in spatial dimension, which is interpolating the learned positional embedding $P \in \mathbb{R}^{n_{h} \cdot n_{w} \times d}$ to $\mathbb{R}^{(n_{h} \cdot n_{w} \times n_{t}) \times d}$, (3-3) interpolating in embedding dimension, which is interpolating the learned postional embedding $P_{image}$ to $P \in \mathbb{R}^{n_{h} \cdot n_{w} \times (d \times n_{t})}$ and reshaping, and (3-4) random initialization.

\subsection{VideoMamba Encoder Block}
\label{section:encoder}
After the pre-processing steps mentioned in Sec. \ref{section:tokenizer} and Sec. \ref{section:embedding}, we obtain $n_{t} \cdot n_{h} \cdot n_{w}$ video tokens. The video tokens are prepended with a class token, and fed into the encoder of our proposed VideoMamba, which comprises $L$ encoder blocks. The architectural design of VideoMamba's encoder block adopts the design of previous works \cite{gu2023mamba, zhu2024vision, liu2024vmamba}, incorporating layer normalization, 1D convolution, and SSM block.
For the SSM block, we propose spatio-temporal forward and backward SSM, which is described in the next section.

\subsection{Spatio-Temporal Forward and Backward SSM} 
\label{section:scanning}

To effectively handle the non-sequential spatial information in VideoMamba encoder block, we process $n_{t} \cdot n_{h} \cdot n_{w}$ tokens through both forward and backward directions.
Since temporal information is inherently ordered, The forward direction can be straightforwardly defined as flattening $n_{t} \cdot n_{h} \cdot n_{w}$ tokens. 
On the other hand, in the backward direction, the question arises: should we reverse temporal information as we do with spatial information, or not?
The combination of sequential and non-sequential spatio-temporal information complicates the determination of the appropriate reversing approach.
With this consideration, we explore three distinct methods for defining the backward scanning direction of the spatio-temporal token sequence, as outlined in Fig. \ref{fig:overview}-(b).

\textbf{Spatio-temporal reversal.}
The first approach is a spatio-temporal reversal, which is equivalent to reversing the order of all flattened $n_{t} \cdot n_{h} \cdot n_{w}$ tokens, compared to $n_{t} \cdot n_{h} \cdot n_{w}$ tokens in the forward direction. 
Figure \ref{fig:block} illustrates this process.
This approach has the advantage of preserving the overall order of input between forward and backward in a similar way to images,
akin to handling the video by concatenating $n_{t}$ frames in a columnar direction, creating a vertically long image.

\textbf{Spatial reversal.}
The second method is a spatial reversal, which does not flip all tokens but only reverses each $n_{h} \cdot n_{w}$ set of tokens, maintaining the order along the temporal axis. 
This approach preserves the temporal order of data across both forward and backward paths, giving the model a clear temporal flow in the given video.  

\textbf{Temporal reversal.}
The last method is a temporal reversal, maintaining the order among the $n_{h} \cdot n_{w}$ tokens while reversing their temporal sequence only. This approach can enrich the understanding of the model for the temporal dynamics, by providing inverse event progression without altering the spatial integrity of the frames.

\section{Experiments}
\label{sec:Experiment}

\subsection{Experimental Setup}
\subsubsection{Datasets.}
For action recognition tasks, we adopt Kinetics-400 (K400) \cite{kay2017kinetics}, Something-Something V2 (SSV2) \cite{goyal2017something}, and HMDB51 (HMDB) \cite{kuehne2011hmdb} datasets. The Kinetics-400 dataset contains $\sim$240k training videos and 20k validation videos from 400 human action classes. Something-Something V2 dataset consists of 168.9k training videos and 24.7k validation videos over 174 classes. HMDB51 dataset is relatively smaller than the Kinetics and the SSV2 datasets, containing around 9.5k training videos and 3.5k validation videos from 51 classes. 
In all experiments, the models are trained on the training videos and then evaluated on the validation videos from the above datasets.

\subsubsection{Implementation details.}
By default, we set the number of blocks $L$ to 24. To align with the model sizes of VideoSwin series, we set the hidden state dimension $d$ to 384. For video tokenizer, tubelet size is set to $s_{t}=2, s_{h} = s_{w} = 16$.

We employed the AdamW optimizer for training. A linear warm-up and a cosine decay learning rate scheduler were both used to train the models for a fixed number of epochs, with a batch size of 64. We initialized the backbone of the network with weights pre-trained on ImageNet1K, while randomly initializing the head layers. The initial learning rate is set to 3e-4. Stronger data augmentation techniques were employed, including label smoothing\cite{szegedy2016rethinking}, RandAugment\cite{cubuk2020randaugment}, and random erasing\cite{zhong2020random}.
For inference, we followed the approach described in \cite{tong2022videomae} by using multiple views (crops) of the video. The final prediction score was computed by averaging the scores from each view. 

\textbf{Kinetics.} We sampled 16 frames from each video with a temporal stride of 2 and resized them to 224$\times$224. We used an AdamW optimizer for 30 epochs with a cosine decay learning rate scheduler and 1 epoch of linear warm-up. A batch size of 64 was utilized.

\textbf{SSV2.} Similar to Kinetics, we used 16 frames with a temporal stride of 2 resized to 224$\times$224. AdamW optimizer was used for training for 35 epochs with a learning rate scheduler and warm-up. Batch size and data augmentation, except for reverse augmentation, were consistent with Kinetics.

\textbf{HMDB51.}  We used the same training strategy as Kinetics for HMDB, with 50 epochs for training.

\subsection{Analysis of Model's Dependency on Temporal Consistency}
In this experiment, we aim to assess the video model's understanding of temporal arrangements by reordering the input frames.
This investigation is crucial for the initial expansion and analysis of image models into video understanding.  
For instance, if the model merely treats video as a bundle of images, the results would likely be similar regardless of how the video is reordered.
To do this, the original video frames (ex. indexed 1 through 8) were subjected to various reordering strategies to test their effects. 
Our reordering strategies for this experiment include:

\textbf{(1) Interleaved reordering}:
Implementing a pattern that consistently alternates between distant frames, this method was designed to test the model's robustness to extreme disruptions in narrative flow.
For instance, the sequence could follow the order $[1\rightarrow 8\rightarrow 2\rightarrow 7\rightarrow 3\rightarrow 6\rightarrow 4\rightarrow 5]$.
\textbf{(2) Pairwise reordering}: In this approach, the frames were reordered in pairs, specifically creating the sequence $[1\rightarrow 2\rightarrow 7\rightarrow 8\rightarrow 5\rightarrow 6\rightarrow 3\rightarrow 4]$ by grouping frames in twos. This was done to ensure that, in scenarios where the tubelet size is set to 2, feature generation would not encounter significant issues.
\textbf{(3) Block-wise reordering}:
This strategy involved reorganizing the video frames into blocks, each comprising half of consecutive frames. 
For example, the sequence $[5\rightarrow 6\rightarrow 7\rightarrow 8\rightarrow 1\rightarrow 2\rightarrow 3\rightarrow 4]$ is created. 

The experimental result on the HMDB dataset is presented in Table \ref{tab: Temporal Ordering}.
The result demonstrates a clear relationship between the complexity of temporal rearrangement and the performance of our model.
The model achieves the best performance when processing videos in their original sequential order, showing its reliance on the inherent temporal flow for optimal understanding.
As the temporal disruption became more severe, from blockwise reordering to pairwise reordering and then to the most distracting interleaved reordering, the performance of the model is progressively declined.
These results show that our model interprets the action in video by reflecting their temporal order.

\subsection{Spatio-Temporal Forward and Backward SSM}
As mentioned in Section \ref{section:scanning}, we compare three different methods for spatio-temporal backward scanning: (1) spatial reversal, (2) temporal reversal, and (3) spatio-temporal reversal. Table \ref{tab:scanning} shows the experimental result on SSV2 and HMDB datasets. 
We did not use an additional class token in this experiment, since the position of the class token can affect the performance. 
The results show that spatio-temporal reversal is the most effective method for backward SSM, emphasizing the importance of the complementary relationship of token orders in forward and backward SSM.
In contrast, spatial reversal is shown to be less advantageous, as it keeps the relative positions of most tokens the same in both the forward and backward paths, impeding the model from fully benefiting from the advantages of bidirectional scanning. Therefore, we select spatio-temporal reversal as our backward scanning method.

\begin{figure}[t]
\begin{minipage}{\linewidth}

\begin{minipage}[t]{0.44\linewidth}
\captionof{table}{Experimental result of VideoMamba's dependency on temporal consistency. We report Top-1 accuracy of ImageNet pretrained model on HMDB dataset.\\}
\label{tab: Temporal Ordering}

\begin{tabular}[c]{>{\centering\arraybackslash}m{3.8cm}>{\centering\arraybackslash}m{1.2cm}}

        \shline        
        Reordering strategy & HMDB                   \\ 
        \hline
        Interleaved       & 51.3                   \\
        Pairwise          & 53.5                   \\
        Block-wise       & 56.5                   \\
        \rowcolor{LightCyan}
        Sequential        & \textbf{58.9}          \\ 
        \shline
\end{tabular}
    
\end{minipage}
\hfill
\begin{minipage}[t]{0.5\linewidth}
\captionof{table}{Different backward scanning methods in spatio-temporal forward and backward SSM. We report the Top-1 accuracy on SSV2 and HMDB, using ImageNet pretrained model.\\}

\label{tab:scanning}
\centering
\begin{tabular}[c]{>{\centering\arraybackslash}m{3.4cm}>{\centering\arraybackslash}m{1.2cm}>{\centering\arraybackslash}m{1.2cm}}
    \shline
    Backward scanning   & SSV2          & HMDB      \\ 
    \hline
    Spatial reversal        & 61.9          & 43.3       \\
    Temporal reversal        &   63.3           & 52.9       \\
    \rowcolor{LightCyan}
    Spatio-Temporal reversal & 64.7 & \textbf{55.2}\\ 
    \shline
    \end{tabular}
\end{minipage}

\end{minipage}
\end{figure}

\subsection{Ablational Study}
\begin{figure}[t]
\begin{minipage}[t]{0.55\linewidth}
\captionof{table}{Positional embedding and initialization method from ImageNet pretrained model on SSV2 and HMDB, as stated in Section \ref{section:embedding}.\\
}
\label{tab: ablation_pe}
\centering
\begin{tabular}[c]{>{\centering\arraybackslash}m{1.9cm}>{\centering\arraybackslash}m{2.3cm}>{\centering\arraybackslash}m{1.cm}>
{\centering\arraybackslash}m{1.cm}}
    \shline
    Pos. Emb.         & Initialize           & SSV2    & HMDB   \\
    \hline
    None          & -         & 63.2    & 48.7   \\
    \hline
    Sinusoid  & -     & 63.3    & 47.5   \\
    \hline
    \multirow{4}{*}{\begin{tabular}[c]{@{}c@{}}Learned\end{tabular}}
                 & Random          & 63.4   & 47.9 \\
                 & Interpolate-s & 63.6   & 49.4 \\
                 & Interpolate-d & 63.6   & 51.5 \\
    \rowcolor{LightCyan}
                 & Expanding    & \textbf{63.7} & \textbf{58.9} \\
    \shline
    \end{tabular}
\end{minipage}
\hfill
\begin{minipage}[t]{0.4\textwidth}
\captionof{table}{Adding regularization on HMDB dataset. Random augmentation \cite{cubuk2020randaugment} and label smoothing \cite{szegedy2016rethinking} are progressively added on K400 initialization.\\} 
\label{tab: ablation_regularization}
\centering
\begin{tabular}[c]{>{\arraybackslash}m{3.5cm}>{\centering\arraybackslash}m{1.2cm}}
\shline
                           & HMDB  \\ 
\hline
ImageNet init.             & 53.6   \\
K400 init.     & 66.6   \\
$+$ Random aug.\cite{cubuk2020randaugment}         & 67.6   \\
$+$ Label smoothing \cite{szegedy2016rethinking}       & 68.6   \\ 
\shline
\end{tabular}
\end{minipage}
\end{figure}

\begin{figure}[t]
\begin{minipage}{\linewidth}
\begin{minipage}[t]{0.44\textwidth}
\captionof{table}{Number of input frames on SSV2 and HMDB dataset, fine-tuning from ImageNet pretrained model.\\}
\label{tab: ablation_num_frames}
\centering
\begin{tabular}[c]{>{\centering\arraybackslash}m{1.2cm}>{\centering\arraybackslash}m{1.2cm}>{\centering\arraybackslash}m{1.2cm}>{\centering\arraybackslash}m{1.2cm}}
        \shline
        Frames & GFLOPs    & SSV2      & HMDB   \\ 
        \hline
        8            & 17.1      & 61.0      & 52.7    \\ 
        16           & 34.2      & 63.7      & 58.9    \\ 
        \rowcolor{LightCyan}
        32           & 68.5      &\textbf{64.2} & \textbf{59.3}  \\ 
        \shline
        \end{tabular}
\end{minipage}
\hfill
\begin{minipage}[t]{0.52\textwidth}
\captionof{table}{Embedding dimension of model on K400, SSV2, HMDB dataset. We used ImageNet pretrained model, and Top-1 accuracy is reported.\\}
\label{tab: ablation_model dim}
\centering
\begin{tabular}[c]{>{\centering\arraybackslash}m{1.2cm}>{\centering\arraybackslash}m{1.2cm}>{\centering\arraybackslash}m{1.2cm}>
{\centering\arraybackslash}m{1.2cm}>{\centering\arraybackslash}m{1.2cm}}
\shline
Dim. & GFLOPs   & K400    & SSV2   &  HMDB  \\
\hline
192       & 8.8        & 69.5       & 54.6    &   56.5     \\
384       & 34.2        & 76.2    & 63.7   & 68.6   \\

\shline
\end{tabular}
\end{minipage}

\end{minipage}
\end{figure}

\subsubsection{Positional embedding.}
In this section, we investigate the impact of different positional embedding strategies (Sec. \ref{section:embedding}) and their initialization methods on the SSV2 and HMDB datasets. As shown in Tab. \ref{tab: ablation_pe}, Omitting positional embeddings leads to lower performance, highlighting their importance. Using sinusoidal embeddings provides a slight improvement, but using learnable positional embedding with appropriate initialization yields better performance. Notably, the initialization by temporal expanding method stands out, achieving the highest accuracy across both datasets, as shown in our table. 
This shows the effectiveness of appropriately initialized positional embeddings from the image model in enhancing the model's ability to process video content, giving the model additional spatiotemporal information.

\subsubsection{Adding regularization.}
As a pioneering work in video recognition model based on Mamba, we conducted experiments to discover training recipes for efficient learning of the model. We sequentially evaluated the impact of various regularization techniques on training on HMDB in Tab. \ref{tab: ablation_regularization}. Using Kinetics-400 initialization considerably increased the accuracy, illustrating the benefits of domain-specific pretraining. Using random augmentation and label smoothing also lead to a slight improvement, each resulting in an improvement of 1\%. This progression demonstrates regularization methods as random augment and label smoothing are effective in training the VideoMamba model on small datasets as well.

\subsubsection{Number of frames.}
We conducted a comparative analysis of the model's performance across varying numbers of frames, focusing on the impact of frame count on accuracy in Tab. \ref{tab: ablation_num_frames}.
It was observed that the model delivered its most robust performance with the longest input sequence of 32 frames. 
This outcome underscores the model's ability to effectively process long-range data, achieving superior performance with linear computational complexity. 
This efficiency showcases the model's adeptness in handling extensive temporal information, a significant advantage over the quadratic complexity typically associated with self-attention mechanisms.

\subsubsection{Embedding dim}
We perform an ablation study on the embedding dimension of the model on K400, SSV2, and HMDB datasets. We compared two different embedding dimension sizes, 192 and 384, and GFLOPs and Top-1 accuracy are reported in Tab. \ref{tab: ablation_model dim}. 
The results clearly show that even in very lightweight scenarios, the model performs well, while a larger embedding dimension can further enhance the understanding ability of VideoMamba.

\subsection{Comparison to Various Video Models}
\subsubsection{HMDB51}
The performance of VideoMamba on the HMDB51 dataset, provided in Table \ref{tab:hmdb}, demonstrates its superiority not only against traditional transformer models like VideoSwin but also against SSM models such as Mamba-ND and S4ND-ConvNeXt-3D. When pretrained on ImageNet-1K, VideoMamba exhibits decent performance with a Top-1 accuracy of 58.9\% for 16 frames and 59.3\% for 32 frames. Furthermore, when we pretrain VideoMamba on K400, a dataset more aligned with video content, its performance notably increases to a Top-1 accuracy of 68.6\% for 16 frames, outperforming all of the compared models.

Compared to VideoSwin-T, a conventional transformer model, VideoMamba utilizes significantly fewer FLOPs and fewer parameters, surpassing VideoSwin-T by 4.9\%. Furthermore, we also outperform concurrent mamba-based architecture \cite{li2024mamba} and the previous SSM-based method \cite{nguyen2022s4nd}, still with fewer parameters.

\begin{table}[t]
\makeatletter\def\@captype{table}\makeatother\caption{Comparison with previous work on HMDB51. $\ddagger$ denotes results from \cite{li2024mamba} and  $\dagger$ is reproduced number for fair comparison. Magnitudes are Mega ($10^{6}$) for Param. The subscript denotes the trained epoch of the model. “N/A” indicates the numbers are not available for us.}
\centering
\label{tab:hmdb}
\resizebox{\textwidth}{!}{
\begin{tabular}{lccccc}
\shline
Method                              & Backbone    & Pretrain    & Frames     & Param     & Top-1      \\
\hline
I3D\cite{carreira2017quo}           & Inception & IN-1K    & 30       & 25.0      & 49.8       \\
SpeedNet\cite{benaim2020speednet}   &S3D-G       &K400   &64           & 9.0 &48.8  \\
VTHCL\cite{yang2020video}           &SlowOnly-R50 &K400  &32         & 32.0&67.9  \\

MemDPC\cite{han2020memory}          &R-2D3D &K400        &40     & 32.0& 54.5 \\
CVRL\cite{qian2021spatiotemporal}   &SlowOnly-R50 &K400  &32 & 32.0 &49.2 \\

\hline
VideoSwin-T$\dagger$\cite{liu2022video}  & Swin-T       & IN-1K & 32                & 27.9         & 54.4         \\
VideoSwin-T$\dagger$\cite{liu2022video} & Swin-T   & K400         & 32          & 27.9         & 69.9       \\
VideoSwin-S$\ddagger$\cite{liu2022video} & Swin-T        & IN-1K & 32            & 54.0        & 58.1             \\
VideoMAE$_{4800e}$\cite{tong2022videomae} & ViT-B      & -              & 16   & 87.0   & 62.6          \\
VideoMAE$_{4800e}$\cite{tong2022videomae} & ViT-B      & K400              & 16   & 87.0   & 73.3          \\
\hline
S4ND-ConvNeXt-3D$\ddagger$\cite{nguyen2022s4nd}  & ConvNeXt  & IN-1K & 30       & 29.0      & 55.2        \\
Mamba-ND$\ddagger$\cite{li2024mamba}       & Mamba              & IN-1K & 32       & 36.0      & 59.0           \\
\hline
\rowcolor{LightCyan} 
VideoMamba       & Mamba        & IN-1K & 16             & 26.3      & 58.9           \\

\rowcolor{LightCyan}
VideoMamba       & Mamba        & IN-1K & 32     & 26.8      & 59.3          \\
\rowcolor{LightCyan} 
VideoMamba       & Mamba        & K400        & 16            & 26.3     & 68.6         \\
\rowcolor{LightCyan} 
VideoMamba       & Mamba        & K400        & 32            & 26.8     & 75.7         \\
\shline
\end{tabular}}
\end{table}

\begin{table}[t]
\centering
\makeatletter\def\@captype{table}\makeatother\caption{Comparison with previous work on Something-Something V2. “Views” indicates temporal clip $\times$ spatial crop and $\dagger$ is reproduced number. Magnitudes are Mega ($10^{6}$) for Param. The subscript denotes the trained epoch of the model.}
\label{tab: ssv2comparison}
\resizebox{\textwidth}{!}{
\begin{tabular}{lcccrcccc}
\shline
Method         & Backbone  & Pretrain         & Frames & View  & GFLOPs & Param & Top-1   & Top-5   \\ 
\hline
SlowFast\cite{feichtenhofer2019slowfast}      & ResNet101 & K400             & 8$+$32    & 1$\times$3 & 106    & 53.3  & 63.1    & 87.6    \\
TSM-RGB\cite{lin2019tsm}         &  ResNet50         & K400             & 16        & 2$\times$3 & 62     & 42.9  & 63.4    & 88.2   \\
\hline
VideoSwin-T$\dagger$\cite{liu2022video}  & Swin-T   & IN-1K & 32        & 1$\times$3 & 88     & 27.8  & 52.3    & 81.9    \\ 
VideoSwin-T$\dagger$\cite{liu2022video} & Swin-T   & K400        & 32        & 1$\times$3 & 88     & 27.8  & 57.2    & 85.7    \\
TimeSformer\cite{bertasius2021space}     &  ViT-B    & IN-21K    & 8         & 1$\times$3 & 196    & 121.4 & 59.5    & N/A     \\
TimeSformer\cite{bertasius2021space}     &  ViT-L    & IN-21K     & 64        & 1$\times$3 & 5549   & 430.0 & 62.4    & N/A     \\
ViViT FE\cite{arnab2021vivit}        &   ViT-L   & IN-21K$+$K400      & 32        & 4$\times$3 & 995    & N/A   & 65.9    & 89.9    \\
VideoMAE$_{2400e}$\cite{tong2022videomae}     & ViT-S  & -             & 16        & 2$\times$3 & 57     & 22.0  & 66.8    & 90.3    \\ 
\hline
\rowcolor{LightCyan}

VideoMamba     & Mamba       & IN-1K      & 16        & 2$\times$3 & 34     &  26.3     & 63.7    & 87.8    \\
\rowcolor{LightCyan}
VideoMamba     & Mamba       & IN-1K      & 32        & 2$\times$3 & 68     &  26.8     & 64.2    &88.0     \\
\shline
\end{tabular}
}
\end{table}


\subsubsection{Something-Something V2}
Table \ref{tab: ssv2comparison} shows that our VideoMamba achieves high performance with reduced computational needs on Something-Something V2.
It outperforms TimeSformer, which has a significantly larger number of parameters.
While VideoSwin-T requires 88 GFLOPs, VideoMamba efficiently runs on just 34 GFLOPs for 16 frames and 68 GFLOPs for 32 frames, also having fewer parameters (26.4M and 26.8M, respectively).
It achieves Top-1 accuracies of 63.7\% and 64.2\% for 16 and 32 frames, respectively, outperforming VideoSwin-T by 7\% in terms of Top-1 accuracy.
This highlights VideoMamba's superior efficiency and its ability to handle datasets that require detailed interpretation of spatial and temporal dynamics with fewer resources.

\subsubsection{Kinetics-400}
Table \ref{tab:k400} shows that VideoMamba also demonstrates remarkable efficiency on the Kinetics-400 dataset, offering competitive performance with a smaller computational resource. 
In addition to these efficiencies, VideoMamba achieves Top-1 accuracies of 76.1\% and 77.7\% for 16 and 32 frames, respectively, demonstrating its capability to process video data effectively with less resource consumption compared to VideoSwin-T and other models of similar size.

\begin{table}[t]
\centering
\makeatletter\def\@captype{table}\makeatother\caption{Comparison with previous work on Kinetics-400. The subscript denotes the trained epoch of the model.}
\label{tab:k400}
\resizebox{\textwidth}{!}{
\begin{tabular}{lcccrcccc}
\shline
Method             & Backbone  & Pretrain      & Frames   & View            & GFLOPs& Param & Top-1 & Top-5\\ \hline
R(2+1)D\cite{tran2018closer} & R(2$+$1)D   & -             & 32         & 10$\times$1   & 75     & 61.8  & 72.0  & 90.0  \\
I3D\cite{carreira2017quo} & Inception          & IN-1K   & 32         & 10$\times$3                & 108    & 25.0  & 72.1  & 90.3  \\
NL I3D\cite{wang2018non}              & ResNet101 & IN-1K   & 32         & 10$\times$3     & 359    & 62.0  & 77.7  & 93.3  \\
SlowFast\cite{feichtenhofer2019slowfast} & R101$+$NL   & -             & 16+64         & 10$\times$3     & 234    & 59.9  & 79.8  & 93.9  \\
\hline
MViT-S\cite{fan2021multiscale}              & MViT-S    & -             & 16         & 5$\times$1      & 33   & 26.1  & 76.0  & 92.1  \\
MViT-B, 16×4\cite{fan2021multiscale}       & MViT-B    & -             & 16         & 5$\times$1      & 71   & 36.6  & 78.4  & 93.5  \\
VideoSwin-T\cite{liu2022video} & Swin-T    & IN-1K   & 32         & 4$\times$3      & 88     & 28.2  & 78.8  & 93.6  \\ 
VideoMAE$_{1600e}$ \cite{tong2022videomae} & ViT-S     & -             & 16         & 5$\times$3      & 57     & 22.0  & 79.0    & 93.8  \\ \hline
\rowcolor{LightCyan}
VideoMamba         & Mamba     & IN-1K   & 16         &  5$\times$3     & 34     & 26.4  & 76.1  & 92.5  \\
\rowcolor{LightCyan}

VideoMamba         & Mamba     & IN-1K   & 32         & 5$\times$3      & 68     & 26.8  &  77.7 & 93.3      \\ 
\shline
\end{tabular}}
\end{table}

\clearpage

\subsection{Analysis of Delta}
\label{sec:delta}
\begin{figure}[t]
    \centering
    \includegraphics[width=\linewidth]{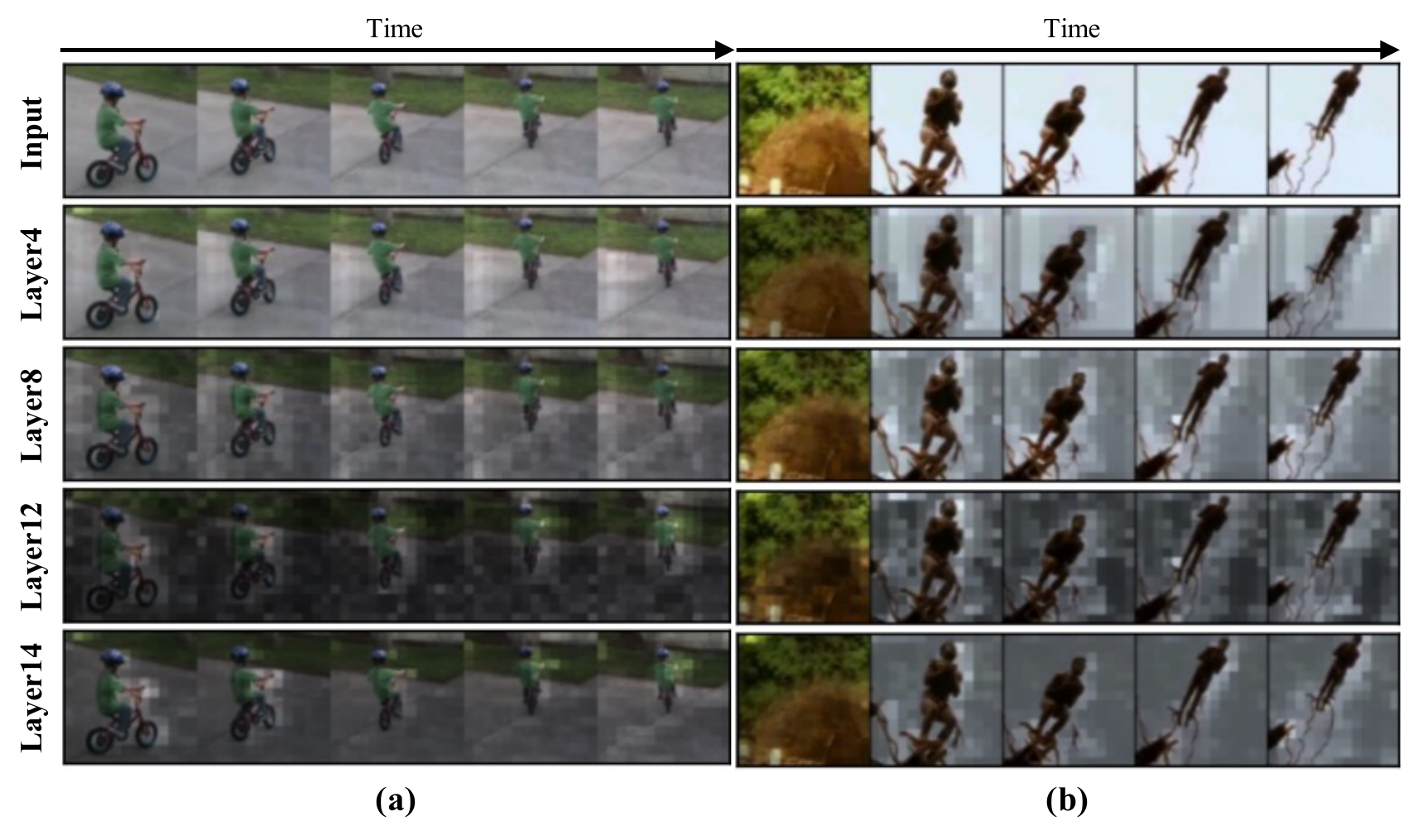}

    \caption{\textbf{Delta visualization.}
Delta plays a crucial role in VideoMamba's context-dependent reasoning by enabling selective emphasis on important aspects when updating the hidden state.
The GT label of (a) is "ride bike," and label of (b) is "dive".
    }
    \label{fig:delta}
\end{figure}

In this section, we explore the significance of $\Delta$ in VideoMamba, highlighting their role in emphasizing the informative features within videos. We specifically examine how $\Delta$ evolves to focus on essential spatio-temporal details over less relevant background noise.
Our visual analysis, presented in Fig. \ref{fig:delta}, demonstrates the selective nature of $\Delta$ in the Spatio-Temporal SSM of VideoMamba. Initially, the model focuses on the overall scene with high $\Delta$ values, using the hidden state's context to distinguish important features. 
As the layer gets deeper, VideoMamba shifts its focus towards dynamic elements, such as motion, by adjusting $\Delta$ values to highlight areas of significant change. For example, increased $\Delta$ values around a moving child in Fig. \ref{fig:delta}-(a) display the model's ability to focus on motion and complex features, rather than static background elements.
Similarly, in Fig. \ref{fig:delta}-(b), $\Delta$ also prioritizes the diving individual while disregarding the background of the initial frame.
This shows how VideoMamba effectively captures long-range dependency with contextual awareness.

\section{Conclusion}
\label{sec:Conclusion}
This study has introduced VideoMamba, an innovative model that marks a significant advancement in video analysis by adapting the pure Mamba architecture to address the complex requirements of video content. Through its utilization of a Spatio-Temporal Selective SSM mechanism, VideoMamba efficiently processes the intricate interplay of spatial and temporal information, achieving a notable balance between computational efficiency and analytical precision. Our extensive evaluations demonstrate VideoMamba's superior performance across various datasets, showcasing its ability to outperform existing models with its adept handling of long-range dependencies and complex video dynamics. 
VideoMamba not only sets new standards in the field but also lays the groundwork for future research, promising to drive significant progress in video recognition and analysis.

{\footnotesize 
\subsubsection{Acknowledgements}
This work was supported by the Institute of Information \& communications Technology Planning \& Evaluation (IITP) grant funded by the Korean government (MSIT) (No.2020-0-00153, Penetration Security Testing of ML Model Vulnerabilities and Defense).
}

\section{Comparison of Computational Complexity}

In this section, we compare the computational complexity of transformer-based model and our VideoMamba.
The multi-head self-attention, which is the basic building block of transformer \cite{vaswani2017attention}, includes computation of the following scaled dot-product attention:
\begin{equation}
\label{eqn:attn}
\mathrm{Attention}(Q, K, V) = \mathrm{softmax}(\frac{QK^{T}}{\sqrt{d_{k}}})V.
\end{equation}

Since the dot-product attention requires calculating $n \times n$ attention matrix, the complexity of self-attention is \textit{quadratic} in input token length $n$.
Therefore, for the input video token with the size ($n_{t} \cdot n_{h} \cdot n_{w}$, $d$), the computation complexity of ViViT \cite{arnab2021vivit}, which uses factorized spatio-temporal encoder, would be $\mathcal{O}((n_{h} \cdot n_{w})^{2} + n_{t}^{2})$.
On the other hand, since our VideoMamba utilizes selective SSM \cite{gu2023mamba}, our model achieves \textit{linear} computational complexity of $\mathcal{O}(n_{h} \cdot n_{w} \cdot n_{t})$.

\section{Additional Delta Visualizations}
\begin{figure}[t!]
    \centering
    \includegraphics[width=\linewidth]{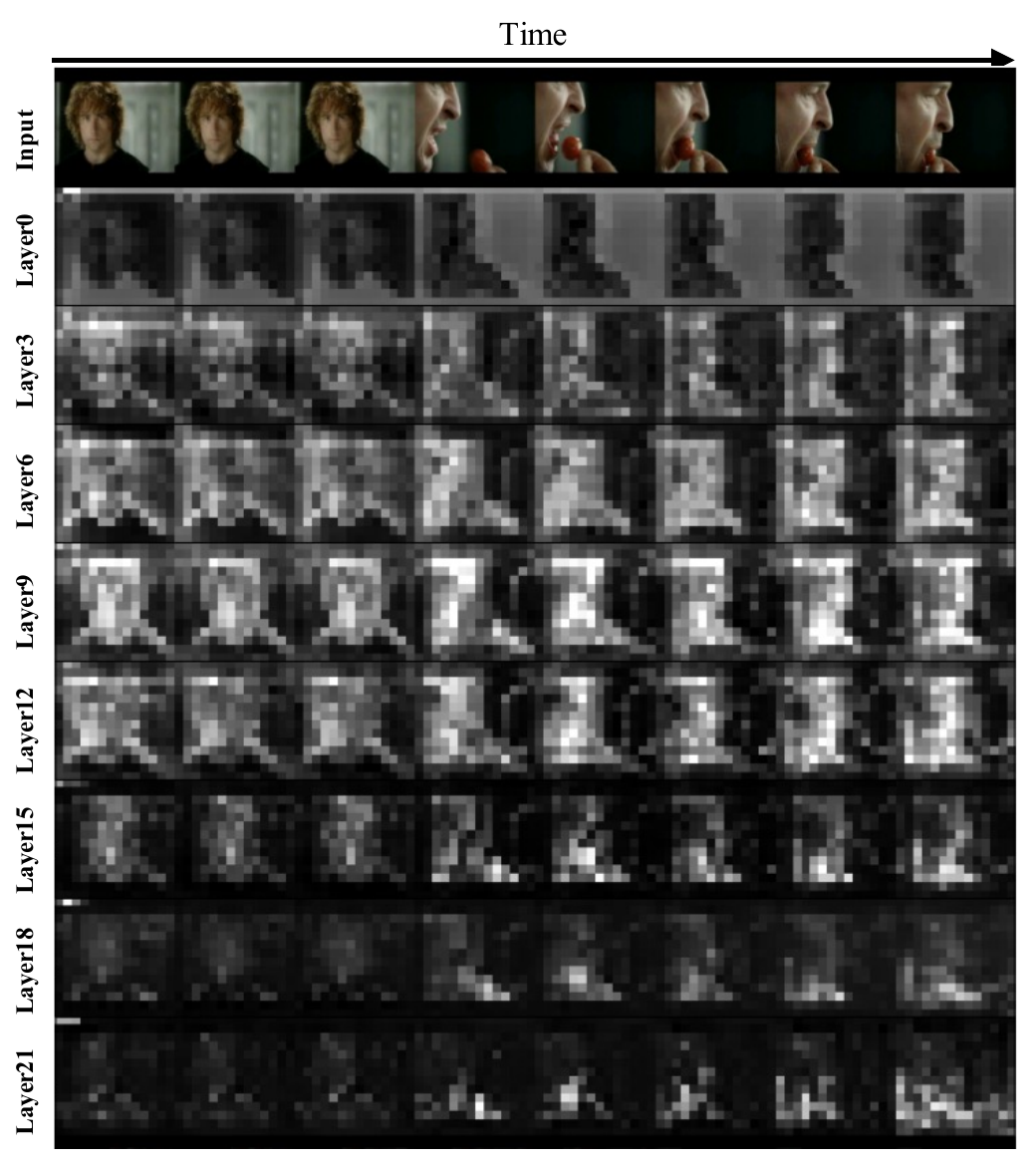}
        \caption{ \textbf{Delta visualizations} on HMDB51 validation set. 
        Darker regions correspond to smaller delta values, indicating the model prioritizing information from previous time steps, and brighter areas represent larger delta values, representing the model placing greater emphasis on the current input. 
        The GT label is ``\textit{Eat}".
        As the network goes deeper into the layers, the delta values decrease, which allows the model to effectively filter out not directly related areas(e.g., upper padded areas) or frames (e.g., initial three frames of layers 18 and 21) and focus on the elements necessary for key parts(e.g., hand and tomato). 
        The high delta values in the initial layers demonstrate the model's process of first understanding the overall scene and then selectively focusing on important details later. 
        Through delta value analysis, we can glean the VideoMamba's capability in performing efficient spatiotemporal reasoning.} 
    \label{fig:delta1}
\end{figure}

\begin{figure}[t!]
    \centering
    \includegraphics[width=\linewidth]{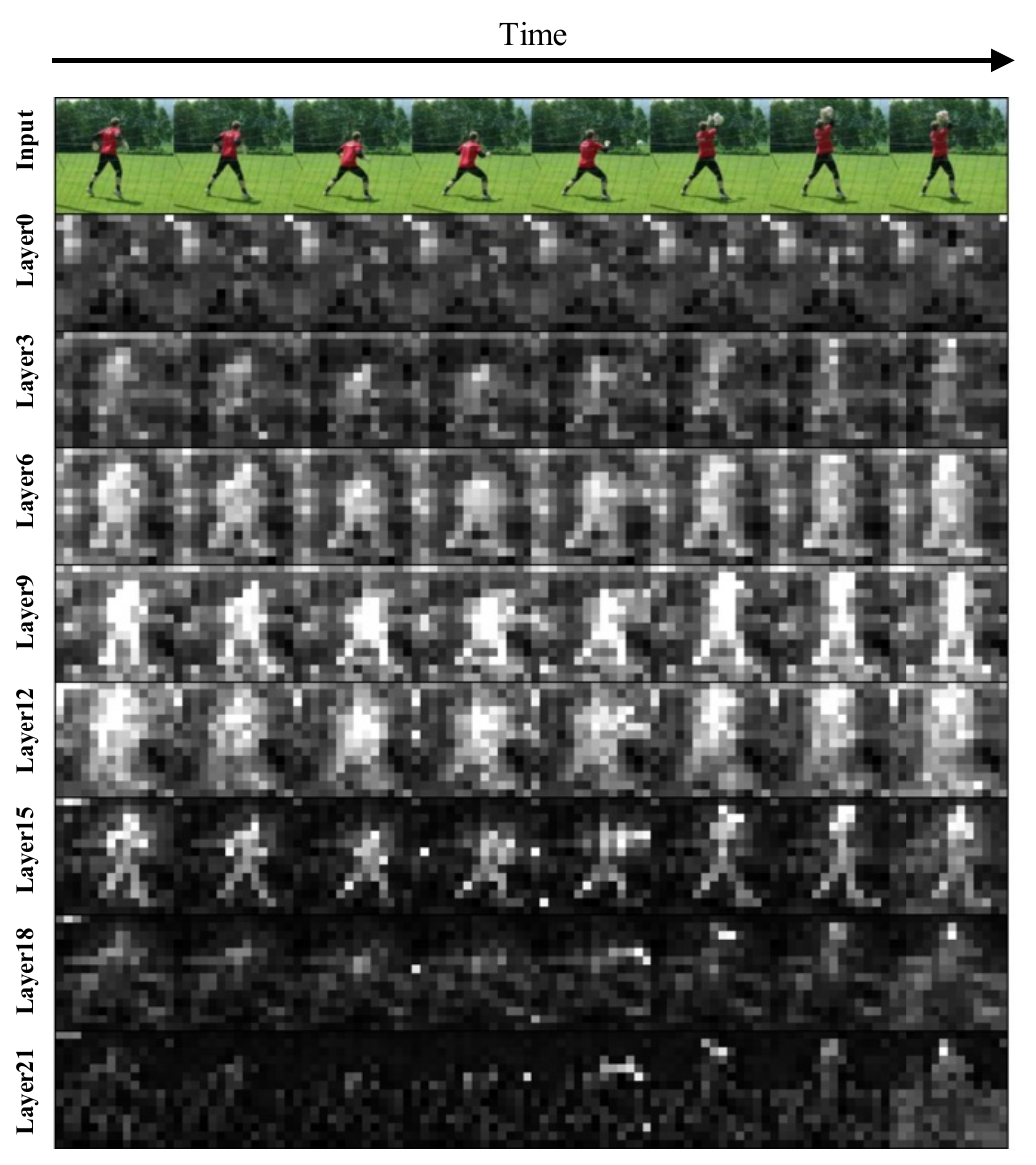}
        \caption{\textbf{Delta visualizations} on HMDB51 validation set. 
        The GT label is ``\textit{Catch}''.
        Within deeper layers, the VideoMamba show a growing emphasis on extracting features relevant to the class of interest (e.g., hand and ball).
        } 
    \label{fig:delta2}
\end{figure}

\begin{figure}[t!]
    \centering
    \includegraphics[width=\linewidth]{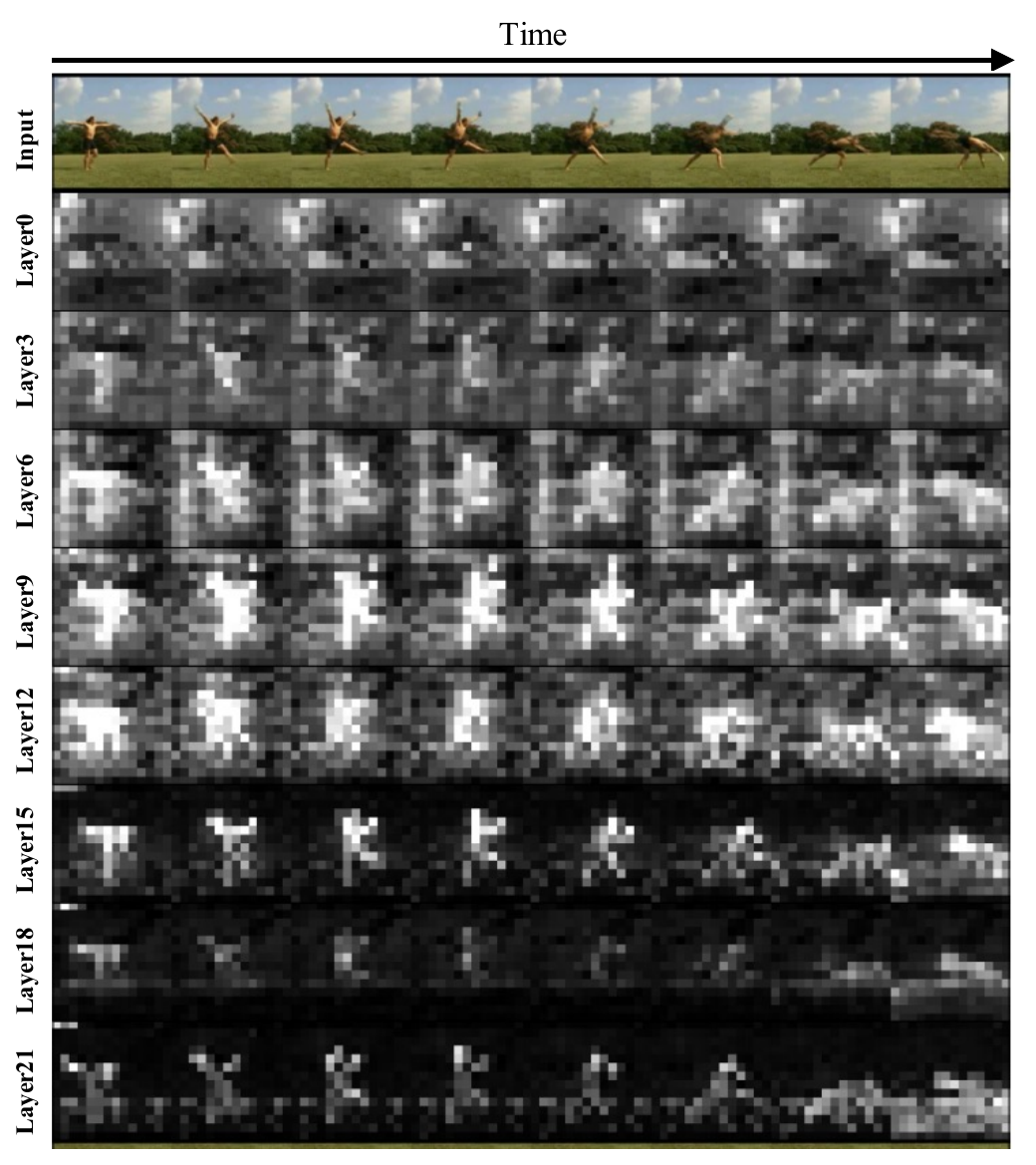}
        \caption{\textbf{Delta visualizations} on HMDB51 validation set. The GT label is ``\textit{Cartwheel}".} 
    \label{fig:delta3}
\end{figure}

To better understand VideoMamba's ability to dynamically select relevant spatio-temporal contexts, we provide additional examples from HMDB51 validation set in \cref{fig:delta1}, \cref{fig:delta2} and \cref{fig:delta3}.
These figures depict the original video sequences alongside their corresponding deltas across multiple layers.

\newpage

\newpage

\clearpage
\newpage
\section{Importance of Pretraining}

\begin{table}[h]
\makeatletter\def\@captype{table}\makeatother\caption{Comparing the effectiveness of ImageNet-1K and Kinetics-400 (K400) pretraining on Something-Something V2 (SSV2) and HMDB51 (HMDB).\\}
\centering
\label{tab:pretrained}
\begin{tabular}{ccc}
\shline
Pretrain       & SSV2          & HMDB      \\
\hline
 $-$           & 44.0        &18.1     \\
ImageNet-1K    & 63.7          & 59.3      \\
 K400          & 63.9      & 68.6      \\

\shline
\end{tabular}

\end{table}

As reported in previous work \cite{bertasius2021space}, the performance of video recognition model depends considerably on pretraining.
In our main experiments, we initialized our models with ImageNet pretrained weights.
In this section, using our base model, we compare the effect of different pretraining datasets on performance, including the experimental results trained from scratch.

Table \ref{tab:pretrained} reports the Top-1 accuracy of differently pretrained models, on Something-Something V2 \cite{goyal2017something} and HMDB51 \cite{kuehne2011hmdb} datasets.
When training from scratch, we trained the model for 100 epochs in Something-Something V2, and 200 epochs in HMDB51.
We observe that training VideoMamba from scratch results in much lower accuracy, especially in small dataset such as HMDB.
We also observe that pretraining on K400 leads to superior performance in HMDB, and slight improvement on SSV2. 

\clearpage

\section{Architecture}
\subsection{Architectural Details}
\begin{table}[h]
\centering
\caption{Architectural Details of VideoMamba.}
\label{tab:arc}
\begin{tabular}{|c|c|c|}
\shline
stage     & VideoMamba                                                             & Output Sizes    \\ \shline
data      & stride 2$\times$1$\times$1 on K400                                      & \textcolor{violet}{3}$\times$32$\times$224$\times$224 \\ \hline
tubelet   & \begin{tabular}[c]{@{}c@{}}Conv3d 2$\times$16$\times$16, \textcolor{violet}{384},\\stride 2$\times$16$\times$16\end{tabular} & 3136$\times$\textcolor{violet}{384}       \\ \hline
encoder   & \begin{tabular}[c]{@{}c@{}}$\left[\begin{array}{c} \text{linear } \textcolor{violet}{384} \rightarrow \textcolor{violet}{384}\times 2\\ \text{conv1d } \textcolor{violet}{384}\times 2 \rightarrow \textcolor{violet}{384}\times 2\\ \text{ST-SSM } \textcolor{violet}{384}\times 2 \rightarrow \textcolor{violet}{384}\times 2\\ \text{linear } \textcolor{violet}{384}\times 2 \rightarrow \textcolor{violet}{384} \end{array}\right] \times 24$\end{tabular} & 3136$\times$\textcolor{violet}{384}       \\ \hline
projector & linear \textcolor{violet}{384} $\rightarrow$ 400 (\# labels)              & 400       \\ \shline
\end{tabular}
\end{table}

Table \ref{tab:arc} outlines the structure of the VideoMamba model, highlighting its various stages from data input to the final projection. The dimensions are emphasized in \textcolor{violet}{violet}.

\textbf{Data Stage}: The initial stage involves processing video data with a stride of 2×1×1 on the K400 dataset. The output size is 3×32×224×224, indicating the transformation of video frames into a tensor with 3 channels (color depth), 32 frames per sequence, and a spatial dimension of 224×224 pixels per frame.

\textbf{Tubelet Stage}: At this stage, a tubelet operation is applied 3d conv with a kernel and stride of 2×16×16, producing an output with a dimension of 3136×384. This illustrates the extracted spatial-temporal features from the input video frames, where 384 represents the feature vector length for each of the 3136 tubelets.

\textbf{Encoder Stage}: This stage, which is repeated 24 times as indicated, involves a sequence of operations starting with a linear transformation from 384 to 384×2 (doubling the feature dimension), followed by a 1D convolution that maintains the feature dimension at 384×2. Spatio-Temporal SSM (ST-SSM) processes these features without altering the dimension, leading to a final linear transformation that maps the features back to a dimension of 384. The output retains the format of 3136×384, emphasizing the consistency in the model's internal representation of features.

\textbf{Projector Stage}: The final stage involves a linear transformation from a 384-dimensional feature vector to the number of labels required for classification. This stage is for adapting the model's learned representations to specific tasks, such as video classification or action recognition. The number of labels is dependent on the application and, thus 400 for K400 dataset in the table.

\clearpage

\subsection{Positional Embedding}
\begin{figure}[h!]
    \centering
    \includegraphics[width=\linewidth]{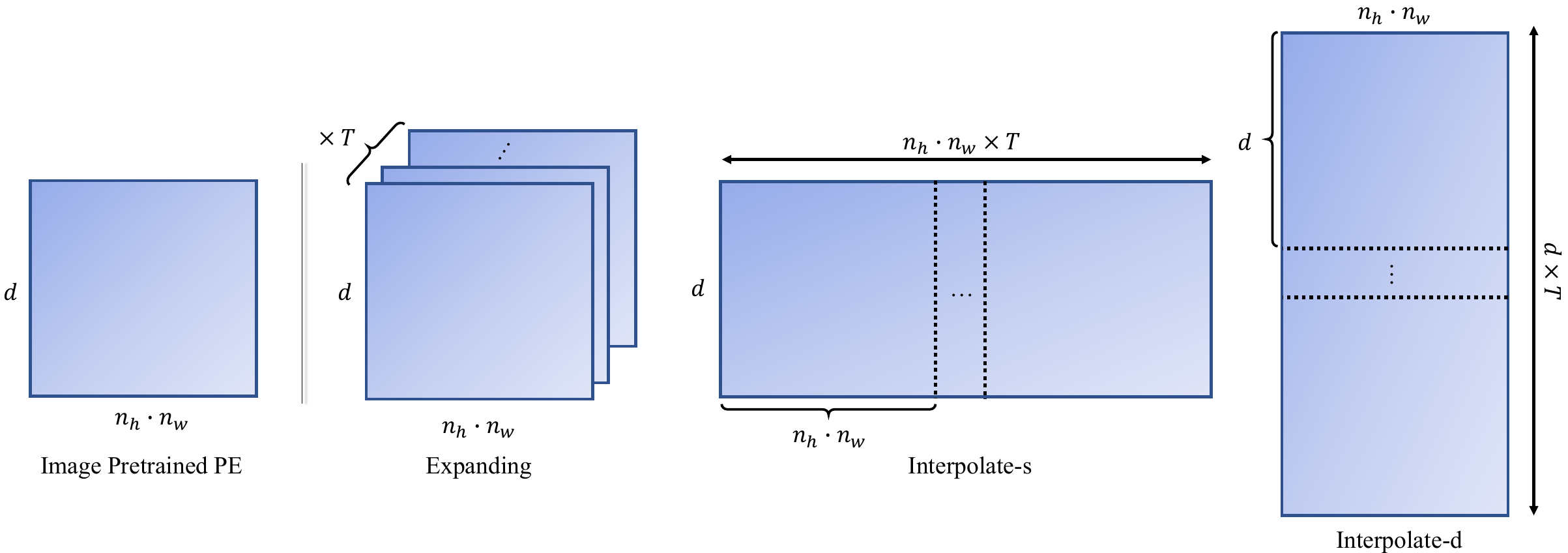}
        \caption{Initialization strategies for learnable positional embeddings in VideoMamba. The figure illustrates three different approaches to modify pretrained image positional embeddings (\(P_{image}\)) for use in video data, which introduces an additional temporal dimension (\(T\)).} 
    \label{fig:pe}
\end{figure}
When employing learnable positional embeddings for video data, initialization plays an important role in leveraging pre-trained image model knowledge. 
Unlike expanding, which technically generates replicated and discontinuous PE for each frame, spatial interpolation (similar to image interpolation) can generate continuous PE across frames. 
We assumed this continuously initialized PE might provide additional temporal information for the model. 
Figure \ref{fig:pe} illustrates the methods we propose \textbf{for adapting \(P_{image}\) to video data.} We propose several initialization techniques for the learnable positional embedding \(P\), starting from the pretrained image positional embedding \(P_{image} \in \mathbb{R}^{n_h \cdot n_w \times d}\), which represents the special case of \(T=1\). Our proposed methods include:

  \textbf{Temporal Expansion:} We replicate \(P_{image}\) along the temporal dimension \(n_t\) times, effectively copying the spatial embeddings across the additional time frames.
  
  \textbf{Spatial Interpolation:} We interpolate \(P_{image}\) in the spatial dimensions to obtain embeddings in \( \mathbb{R}^{(n_h \cdot n_w \times n_t) \times d}\), matching the spatial-temporal structure of video data.
  
  \textbf{Embedding Dimension Interpolation:} We interpolate \(P_{image}\) in the embedding dimension, expanding it to \( \mathbb{R}^{n_h \cdot n_w \times (d \times n_t)}\) before reshaping, to integrate temporal information.

Each method is designed to adapt the effective spatial embeddings from image models to the spatio-temporal domain of video data.

\pagebreak
\section{Implementation Details}
\begin{table}[h!]
\makeatletter\def\@captype{table}\makeatother\caption{Training setting for VideoMamba}
\centering
\label{tab:imple_all}

\begin{tabular}{c|ccc}
\shline
config & K400 & SSV2 &HMDB \\
\hline
optimizer               &\multicolumn{3}{c}{\textit{AdamW}} \\
optimizer momentum      &\multicolumn{3}{c}{$\beta_{1},\beta_{2}=0.9,0.999$} \\
weight decay            &\multicolumn{3}{c}{0.05} \\
learning rate schedule  &\multicolumn{3}{c}{\textit{CosineAnealing}} \\
learning rate           &\multicolumn{3}{c}{3e-4}\\
batch size              &\multicolumn{3}{c}{64}\\
warmup epochs           &1&1&5\\
total epochs            & 30&35&50\\
drop path               &\multicolumn{3}{c}{0.1}\\
repeated augmentation   &\multicolumn{3}{c}{ \textit{no}}\\
RandAug \cite{cubuk2020randaugment}             & \multicolumn{3}{c}{(9,0.5)}\\
label smoothing \cite{szegedy2016rethinking}         &\multicolumn{3}{c}{0.1}\\
flip augmentation       &\textit{yes}&\textit{no}& \textit{yes}\\

\shline
\end{tabular}

\end{table}

In this section, we provide additional experimental details.
Table \ref{tab:imple_all} summarizes the hyperparameters employed for all experiments. 
We opted for the AdamW optimizer with cosine learning rate schedule. 
A consistent batch size of 64 was maintained across all experiments. 
For the large-scale Kinetics-400 dataset, we leveraged two NVIDIA A100 GPUs. Conversely, for the smaller Something-Something V2 (SSV2) and HMDB51 datasets, we employed eight NVIDIA 3090 GPUs for training. 
We initialized all models with pre-trained weights \cite{zhu2024vision} obtained from the ImageNet dataset.
We carefully re-implement VideoSwin-T with the VideoSwin-B training strategy \cite{liu2022video} for the SSV2 dataset and the HMDB51 dataset, with the same augmentation strategy as ours.
\clearpage

\section{Inference Speed} 
\begin{figure}[h!]
  \centering
  \includegraphics[width=0.7\linewidth]
  {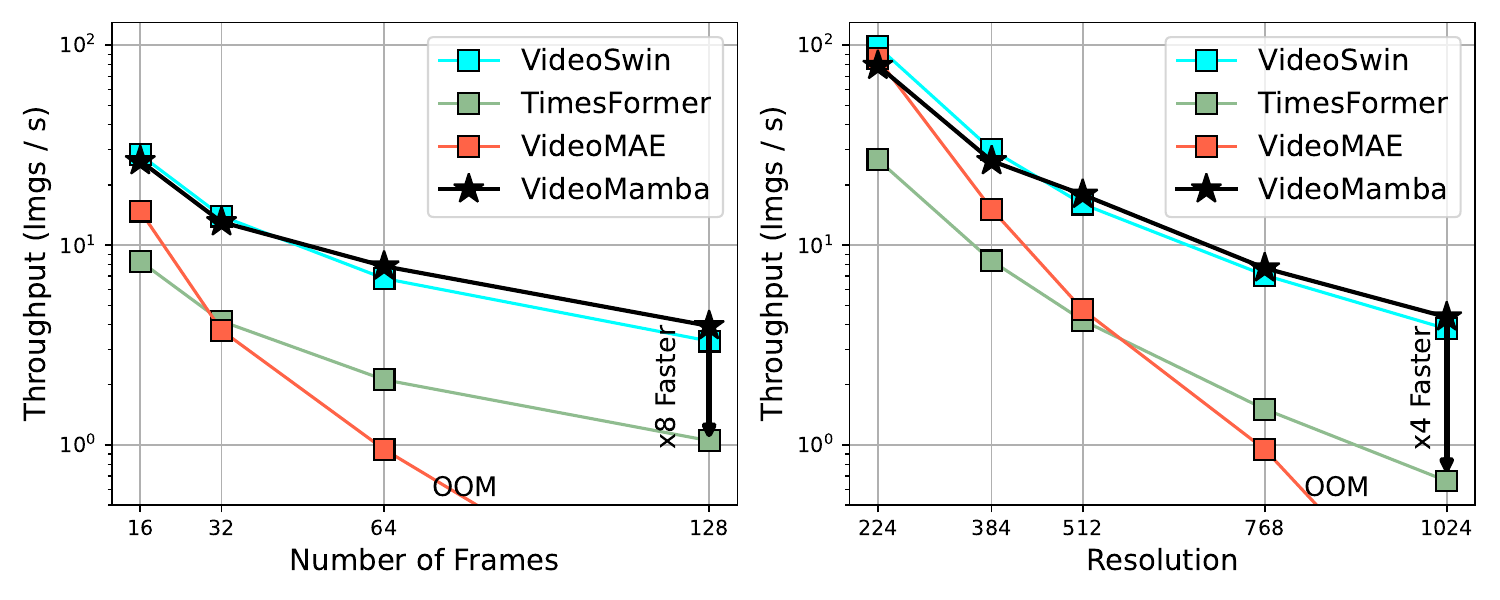}
   \caption{Comparison of Inference
Speed (Throughput).}
   \label{fig:throughput}
\end{figure}
In Fig. \ref{fig:throughput}, our model (black curves)
demonstrates comparable or even faster ($\times 8$) inference speed compared to transformer-based models, especially with longer and higher-resolution videos.

\section{Long-Term Video Modeling} 
To further validate VideoMamba's long-term modeling capabilities, we conducted additional experiments on the Breakfast dataset, which contains longer untrimmed videos. 
Our VideoMamba$_{f64}$ model, using 64 input frames, achieves state-of-the-art performance with 91.5\% accuracy on Breakfast, surpassing all previous methods.

\begin{table}[ht]
\centering
\caption{Long-Term Video Modeling Results on Breakfast.}
\begin{tabularx}{0.6\columnwidth}{lccc}
\shline
Method     & Backbone  & Pretrain & TOP-1  \\ 
\hline
Timeception       & 3D-ResNet  & IN-1K+K400  & 71.3 \\
GHRM              & I3D        & IN-1K+K400  & 75.5 \\
\hline
Distant S. & TimeSformer& IN-21K+HTM  & 89.9 \\
Turbo$_{f32}$      & VideoMAE-B & K400        & 86.8 \\
\hline
ViS4mer$_{f32}$              & Swin-B+SSM     & IN-21K+K600 & 88.2 \\
LSMCL$_{f64}$             & Swin-B+SSM     & K600        & 90.1 \\
\hline
Ours$_{f32}$  & Mamba      & IN-1K+K400  & 90.4 \\
Ours$_{f64}$  & Mamba      & IN-1K+K400  &  \textbf{91.5} \\

\shline
\end{tabularx}
    
    \label{tab:longterm}
    
\end{table}

\clearpage

\section{Applicability to Other Video Tasks} 
In addition to action recognition, 
our additional experiments 
present strong performance in action detection and temporal segmentation. This demonstrates VideoMamba's potential as a versatile and efficient backbone for various video understanding tasks.
\subsubsection{Action Detection.} 
Compared to models with similar size, Tab. \ref{tab: ava} shows VideoMamba outperforms the transformer-based VideoSwin-T backbone on the AVA action detection dataset, while requiring less computation (34 vs 44 GFLOPs).

\subsubsection{Temporal Action Segmentation.} 
Table \ref{tab:tas} shows that integrating VideoMamba into the ASFormer model leads to improved performance on the GTEA dataset, especially in terms of the F1 score and edit score.

\begin{table}[h]
\centering
\caption{Action Detection on Results AVA 2.2.}
\begin{tabularx}{0.6\columnwidth}{lccc}

\shline
Method      & Backbone      
& GFLOPs ($\downarrow$) 
& mAP   \\
\hline
CVRL$_{f32}$      & SlowOnly-R50  
& 42      
& 16.3  \\
VideoMAE$_{f16}$     & ViT-S     
& 57      
& 22.5  \\
VideoSwin$_{f16}$  & Swin-T        
&      44   
& 18.0      \\
\hline
Ours$_{f16}$ & Mamba        
&        \textbf{34} 
& \textbf{22.1}      \\

\shline
\end{tabularx}

\label{tab: ava}
\end{table}

\begin{table}[h]
\centering

\caption{Action Segmentation Results on GTEA.}
\begin{tabularx}{0.6\columnwidth}{lccccc}
\shline
Method     &  \multicolumn{3}{c}{F1@\{10,25,50\}}  &    MoF &Edit \\ 
\hline
BCN        &  88.5   &  87.1  &  77.3 &  79.8  &84.4\\
MS-TCN++   &   88.8  &  85.7  &  76.0 &  \textbf{80.1}  &83.5\\
ASFormer   &   90.1  &  88.8  &  79.2 &  79.7  &84.6\\
\hline
ASFormer w/ Ours  &  \textbf{90.6} & \textbf{89.7}  & \textbf{79.9} &79.6&\textbf{86.6} \\
\shline
\end{tabularx}
    \label{tab:tas}
\end{table}

\clearpage
%
%

\bibliographystyle{splncs04}
\bibliography{egbib}

\end{document}